\documentclass[letterpaper, 10 pt, conference]{ieeeconf} 

\IEEEoverridecommandlockouts

\usepackage{graphicx} % for pdf, bitmapped graphics files
\usepackage[utf8]{inputenc}
\usepackage[hidelinks]{hyperref} % to have hyperlinks and for the \ref macros
\usepackage{subcaption}
\let\labelindent\relax
\usepackage[inline]{enumitem} % inline enumerate
\usepackage[capitalise,nameinlink]{cleveref} % for easier referencing

\def\ie{\textit{i.e.,}}
\def\eg{\textit{e.g.,}}
\def\etal{\textit{et al.}}

\title{\LARGE \bf%
On the Potential of Smarter Multi-layer Maps}

\author{Francesco~Verdoja and Ville~Kyrki%
\thanks{Presented at the \emph{Perception, Action, Learning (PAL)} workshop at
ICRA 2020. This work was supported by the Strategic Research Council at Academy
of Finland, decision 314180.}
\thanks{F.~Verdoja and V.~Kyrki are with School of Electrical Engineering, Aalto
University, Finland. 
\texttt{\{first.surname\}{@}aalto.fi}}}

\begin{document}

\maketitle
\thispagestyle{empty}
\pagestyle{empty}

%%%%%%%%%%%%%%%%%%%%%%%%%%%%%%%%%%%%%%%%%%%%%%%%%%%%%%%%%%%%%%%%%%%%%%%%%%%%%%%%

\begin{abstract}
    The most common way for robots to handle environmental information is by
    using maps. At present, each kind of data is hosted on a separate map, which
    complicates planning because a robot attempting to perform a task needs to
    access and process information from many different maps. Also, most often
    correlation among the information contained in maps obtained from different
    sources is not evaluated or exploited.
    
    In this paper, we argue that in robotics a shift from single-source maps to
    a multi-layer mapping formalism has the potential to revolutionize the way
    robots interact with knowledge about their environment. This observation
    stems from the raise in metric-semantic mapping research, but expands to
    include in its formulation also layers containing other information sources,
    \eg{} people flow, room semantic, or environment topology. Such multi-layer
    maps, here named \emph{hypermaps}, not only can ease processing spatial data
    information but they can bring added benefits arising from the interaction
    between maps. We imagine that a new research direction grounded in such
    multi-layer mapping formalism for robots can use artificial intelligence to
    process the information it stores to present to the robot task-specific
    information simplifying planning and bringing us one step closer to
    high-level reasoning in robots.
\end{abstract}

%%%%%%%%%%%%%%%%%%%%%%%%%%%%%%%%%%%%%%%%%%%%%%%%%%%%%%%%%%%%%%%%%%%%%%%%%%%%%%%%

\section{Introduction}
\label{sec:intro}

Environmental awareness is a crucial skill for robotic systems intended to
autonomously navigate and interact with their surroundings. For indoor robots,
2D occupancy grid maps are often used as internal representation of the
environment where a robot is navigating. These maps represent the world by using
a grid of cells each containing the belief of the system over the occupancy of
the area of the world corresponding to each cell. These maps are usually built
automatically by using 2D laser scanners (lidars) mounted on the robot and are
used for localization and path planning in the environment
\cite{siegwart_introduction_2011}. However, these maps suffer from many
limitations, \eg{} 2D lidars are not able to detect transparent obstacles such
as glass, and are limited to measuring occupancy at a single height, incapable
of inferring the true occupancy of complex objects such as tables
\cite{lundell_hallucinating_2018}. These limitations can partially be addressed
by moving to 3D maps, where a representation of the environment is built using
3D lidars or RGB-D cameras and maintained either as a mesh, point-cloud, or
voxel grid \cite{hsiaoKeyframebasedDensePlanar2017}. However, while both these
representations are able to effectively record the static components of an
environment, their ability to represent its dynamic aspects is limited, \eg{}
representing the possibility of a door being either open or closed, objects
changing position or people movement \cite{krajnik_warped_2019}. Also, these
maps contain enough information for a robot to be able to navigate, but are
insufficient for tasks requiring information other than environment
traversability. For example, they do not host any semantic information about the
environment, \ie{} what the object and rooms in the environment are
\cite{missura_minimal_2018}.

In the past few years, advances in computer vision and machine learning have
increased the ability of autonomous agents to understand the world around them.
In particular, semantic interpretation of sensor output is bringing improved
reasoning capabilities and safety in applications like autonomous vehicles and
indoor service robots \cite{mozos_categorization_2012,bojarski_end_2016}.
Consequentely, the number of techniques proposing to create maps that fuse this
growing amount of richer information on metric maps is increasing
\cite{xiaGibsonEnvRealWorld2018,rosinolKimeraOpenSourceLibrary2020}.

The ability for service robots to exploit these advancements is crucial for
their widespread adoption, as their ability to have a deeper understanding of
their environment will be required for them to solve everyday tasks, from
interacting naturally with the user, to fetching objects in the environment.

\begin{figure}
	\centering
	\includegraphics[width=\linewidth]{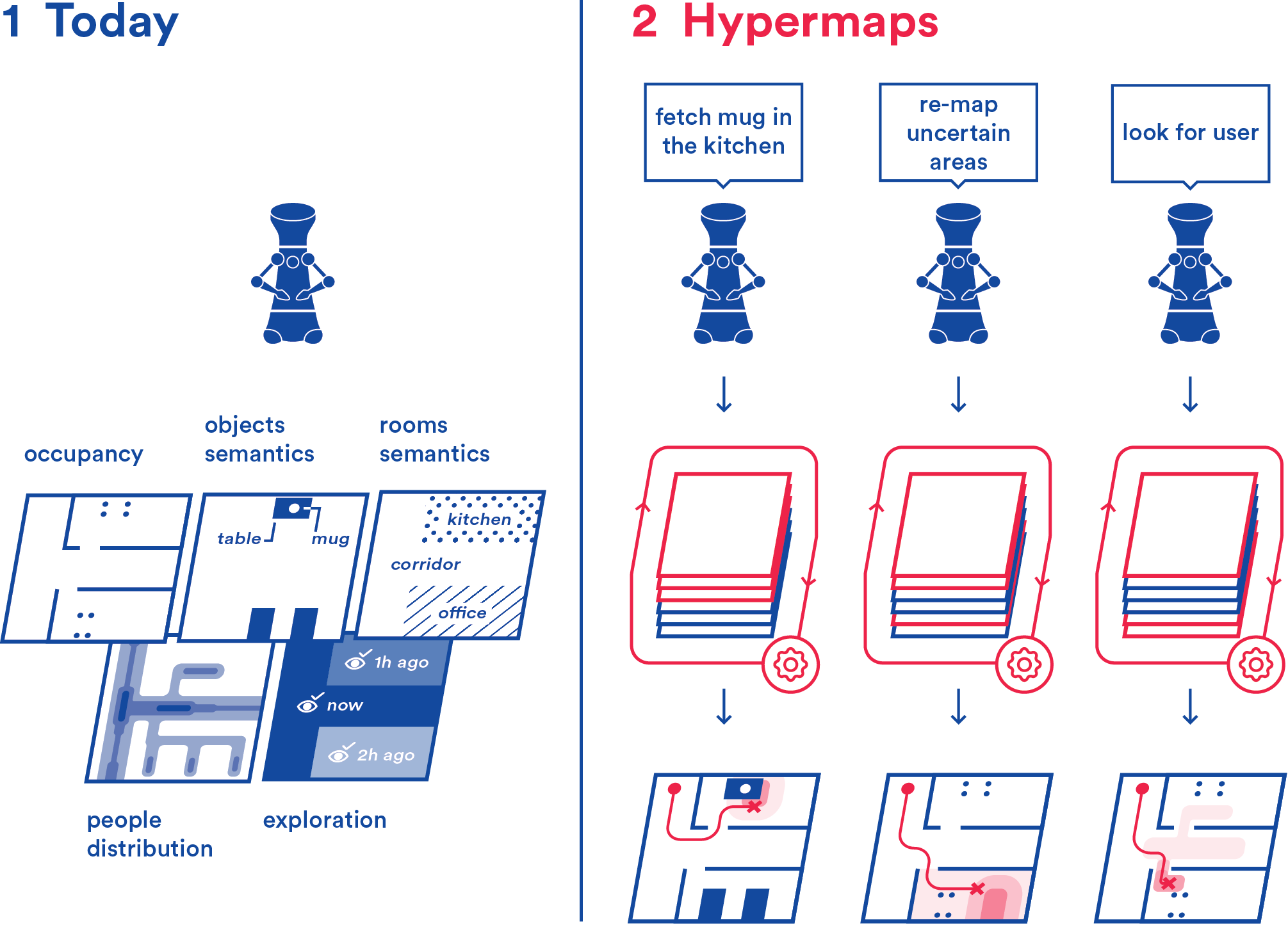}
	\caption{\label{fig:hypermaps}The proposed mapping shift: from single-source
	maps, to a multi-layer mapping framework, an \emph{hypermap}, which
	processes the mapping data to the benefit of the high-level agent.}	
\end{figure}

At the current state of technology, however, any new piece of information the
robot needs to record about the environment is often stored in a separate map:
an occupancy map for navigation, a semantic map for object localization, a map
tracking people movement, and so on.

An example of these more high-level tasks is autonomous semantic exploration,
where a robot's task is to label objects in the environment it is in while
autonomously navigating in it. To solve this task, the inteligent agent guiding
the robot needs to maintain at least an occupancy map of the environment for
navigation, an exploration map keeping track of which areas have already been
visited, and a semantic map where locations and labels of objects in the
environment are recorded.

Each of these maps usually has different implementations and is handled by the
agent independently from the others. This approach has several shortcomings:
\begin{itemize}
	\item The agent needs to interact with the different map implementations,
	which as the number of maps required for advanced applications grows, can
	become a burden. This is particularly challenging for learning agents, for
	which richer and diverse inputs might complicate architecture design and
	training.
	\item The automatic map generation algorithms used for populating these maps
	are prone to errors. This is most critical when maps are populated by
	state-of-the-art deep learning algorithms as most of the times these
	techniques provide point estimates and are unable to measure the confidence
	of their prediction. 
	\item Treating each map independently is suboptimal, as additional
	information could be extracted by observing the interaction and correlation
	amongst different maps.
\end{itemize}

Recent works are starting to combine semantic and metric information in single
maps, by assigning a semantic label to each metric location in the environment
\cite{xiaGibsonEnvRealWorld2018,rosinolKimeraOpenSourceLibrary2020}. However, in
this paper we argue that to address these shortcomings and exploit all the
information maps provide, we need to go beyond semantic-metric maps. We argue
that a paradigm shift from single-source maps to a multi-layer mapping
formalization---which we will call an \emph{hypermap}---is necessary. In such a
framework, each different kind of map information is maintained in relationship
with each other and, moreover, AI techniques could be used to extract additional
knowledge arising from the layer's relationships as well as to simplify the
high-level agent's interaction with the data stored in the maps. A graphical
example of these interactions is shown in \cref{fig:hypermaps}. In the rest of
this article, we will discuss the benefits such a paradigm shift could bring to
robot autonomy as well as possible interesting research directions that it would
enable.

\section{Related works}
\label{sec:related}

\begin{figure}
	\centering
	\includegraphics[width=\linewidth]{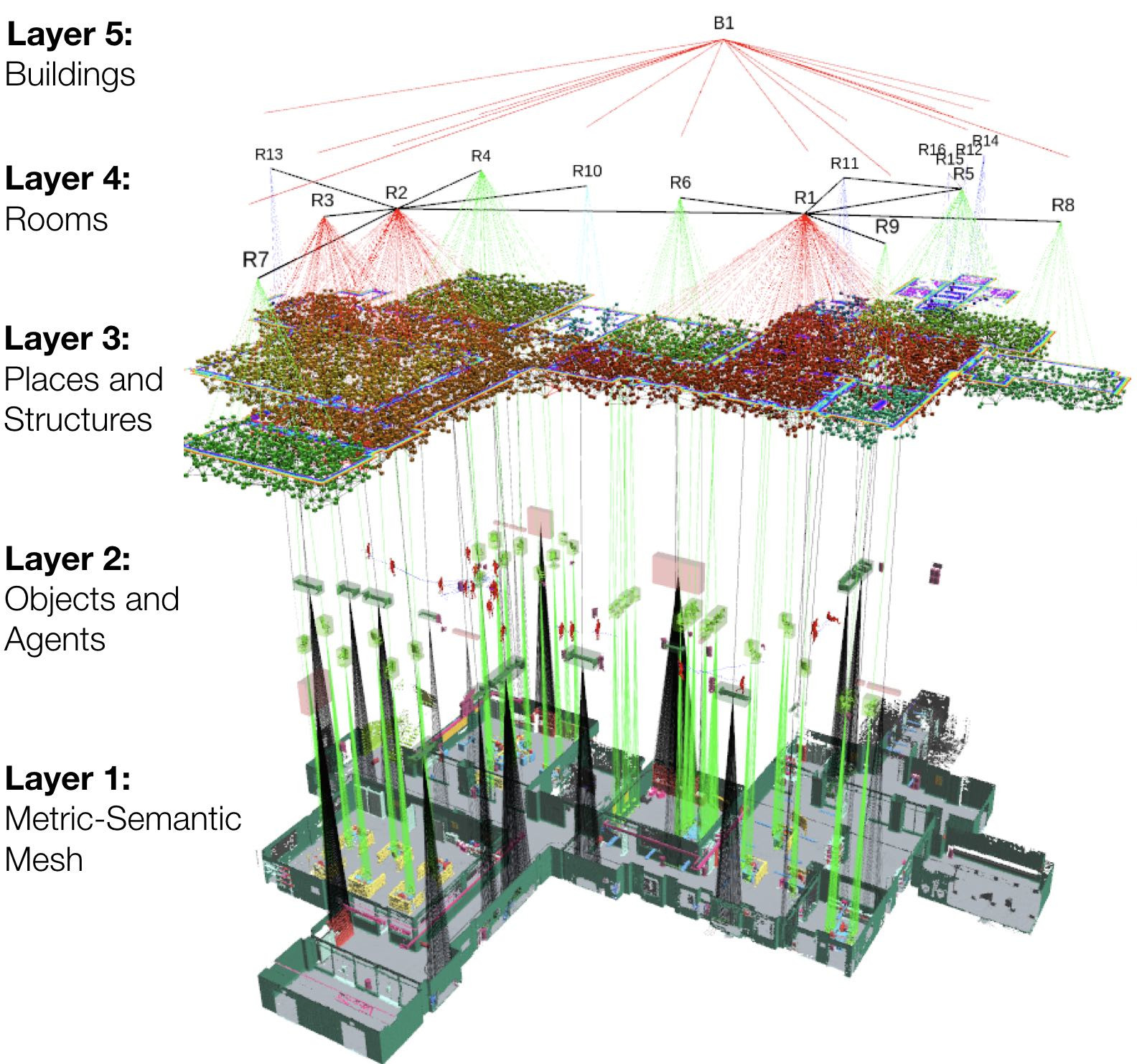}
	\caption{\label{fig:graph}The multi-layer graph mapping formalism proposed
	in \cite{rosinol3DDynamicScene2020}; image courtesy of the original work.}
\end{figure}

While a literature on multi-layer mapping exists, it has so far mostly focused
on hierarchical conceptual abstractions, where each layer captures properties of
the environment at a different granularity: starting from metric, they extract
connectivity graphs and environment topology, to finally construct a spatial
ontology \cite{zenderMultiLayeredConceptualSpatial2007,
zenderConceptualSpatialRepresentations2008}. However, the architectures proposed
in those works rarely include multiple maps capturing the environment as seen by
different sensor modalities but rather abstract the content of the metric map to
facilitate reasoning and human-robot interaction
\cite{crespoRelationalModelRobotic2017}. While these are certainly
caractheristics that should be preserved while building hypermaps, we argue that
they are not suffient and that the modern advancements in computer vision and
machine learning enable us to map a more varied array of properties of the
environment, which might be obtained both from direct sensor reading and virtual
AI-based sensors.

Recently, a couple of multi-layer mapping formalisms have been proposed that
looked at including different sensor modalities. We proposed an early idea for
2D hypermaps in \cite{zaenkerHypermapMappingFramework2019}, where we present a
multi-layer framework composed of a metric, semantic, and exploration layers,
and demonstrate its application in the context of autonomous semantic
exploration. Moreover, the very recent 3D multi-layer graph mapping approach
proposed by Rosinol \etal{} \cite{rosinol3DDynamicScene2020}, fits very well
under the hypermap formalism. In that work, the authors develop a hierarchical
graph (shown in \cref{fig:graph}) connecting the different map abstractions in
each layer and demonstrate the ability of such a system to be used to simulate
realistic dynamic environments. That map comprises 3D metric and semantic
layers, an object layer where a mesh of each individual semantic entity is
maintained in relationship to its position in the environment, as well as
environment abstractions like topological and connectivity graphs.

\section{Proposed research directions}
\label{sec:ai}

At its core, an hypermap centrilizes all the spatial knowledge available to the
robot and maintains a network of relationships between different abstractions
and representations of the environment. Once such a framework is in place, many
interesting possibilities open. By maintaining a network of interactions between
different layers, it is possible to correlate information obtained from
different sources. To give some examples, this could enable us to study the
following: 
\begin{itemize}
	\item \emph{Anomaly detection and correction:} errors in one layer could be
	noticed and corrected by looking at the interaction with other layers. Once
	anomalies are detected, an active strategy can be employed to re-map that
	portion of the environment to eventually correct the anomaly.
	\item \emph{Multi-map knowledge extraction:} many kinds of reasoning could
	benefit from being able to easily access information on multiple layers. As
	an example, occupancy of an area of the environment could be considered
	uncertain after a certain time if the area is marked on the semantic layer
	as a chair, as chairs tend to move. This semantics-based environment
	dynamics could be used to perform safer path planning by avoiding areas of
	high uncertainty.
	\item \emph{Content estimation:} when only a part of the environment has
	been mapped in a specific map, information from the other layers about the
	rest of the environment could be used to estimate what the map content is
	for the unexplored areas, exploiting correlation of the available part of
	the map.
	\item \emph{Virtual planning layers:} Aside for the layers populated by
	sensors (\eg{} occupancy, temperature) and machine learning (\eg{} semantic,
	people movement), new virtual layers could be populated combining
	information from different layers. These layers could be developed in such a
	way to be the ideal planning map for a specific task. As an example, to
	avoid glass walls, a layer combining information from the occupancy layer
	and people movement could be used for a robot to navigate in areas where
	humans navigate, assuming that, even if the robot does not see any obstacle,
	if no human has ever moved through that opening, it is probably impassable.	
\end{itemize}

Some of these problems have parallels in other disciplines, so to tackle these
open research directions, literature from these disciplines can be leveraged and
bridged with the mapping literature. This includes for example, the literatures
on anomaly detection in hyperspectral images
\cite{changAnomalyDetectionClassification2002,verdojaGraphLaplacianImage2020}
and graph signal processing
\cite{narangSignalProcessingTechniques2013,cheungGraphSpectralImage2018}.

\section{Effects on high-level reasoning}
\label{sec:reasoning}

At present, high-level planning involving robot mobility is often posed as a
reinforcement learning (RL) problem where the robot learns a policy to perform a
course of action that leads to the desired goal. The goal is usually tied to the
completion of a task involving some degree of human understanding and
environment navigation.
RL-based solutions to these tasks frequently work without employing any map
\cite{zhuTargetdrivenVisualNavigation2017,chiangLearningNavigationBehaviors2019}
but rather they embed all environmental knowledge inside the policy. This has a
few drawbacks however, namely the difficulty for these policies to generalize to
different environments or tasks, and the inability to guarantee
long-term task completion, since the action selection usually happens on a 
planning horizon shorter than the one required for the full task.

For global long-term task completion, the use of reinforcement learning is
rarer, and more traditional techniques, like A* or rapidly exploring random tree
(RRT), are usually employed. This has the limit that while these methods can
provide good solutions to simple navigation tasks, they are often insuffient for
higher-level planning, involving a richer understanding of the environment.

\begin{figure}
	\centering
	\begin{subfigure}[b]{.482\linewidth}
		{\includegraphics[width=\linewidth]{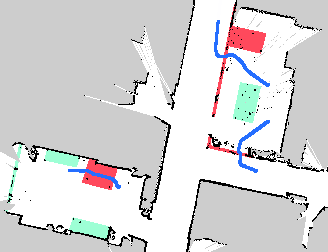}
		\caption{\label{fig:nav_slam}SLAM map}}
	\end{subfigure}
	\begin{subfigure}[b]{.482\linewidth}
		{\includegraphics[width=\linewidth]{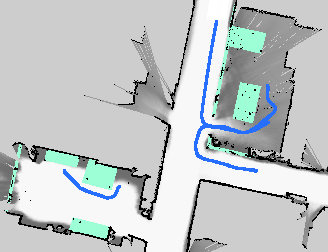}
		\caption{\label{fig:nav_laplace}Uncertainty map}}
	\end{subfigure}
	\caption{\label{fig:nav}Three sample navigation trajectories (in blue) 
	executed on a SLAM map or the uncertainty map presented in
	\cite{verdoja_deep_2019}. The true occupancy of objects invisible to the
	laser is overlaid on the maps, in green when no collision occurred or in red
	in case of collision. The uncertainty map is shown in shades of gray. Image
	courtesy of the original work.}
\end{figure}

We argue that one underobserved aspect that could improve planning performance
is processing the maps themself. In the previous section, we already
hinted at how hypermaps could be used to generate task-specific maps to
simplify planning. As an early example of that, in \cite{verdoja_deep_2019}, we 
presented how planning on virtual layers can enable the robot to perform
safer path-planning: we showed a robot avoiding obstacles---normally invisible
on a 2D occupancy map---by planning with a very simple D* policy on a virtual
map including deep network uncertainty over obstacle distances (\cref{fig:nav}).

Some recent work showed how the integration of traditional sampling-based path
planning with a deep RL planner could effectively improve high-level task
completion in different scenarios \cite{faustPRMRLLongrangeRobotic2018}. In that
work, the map used was a simple metric map and the task constraints were encoded
in the RL planner, but we imagine that similar approaches on task-specific maps
built by combining information coming from multiple layers could enable to
transfer policies between different environments and tasks more easily by
changing the underlying map.

\section{Conclusions}
\label{sec:concl}

In this paper we argue that in robotics there is the need to move from
single-source maps to multi-layer ones---which we named \emph{hypermaps}---built
by combining information coming from different sensors and AI-based sources and
maintaining their inter-layer relationships and correlations. Recent methods
performing semantic-metric mapping are a first step in that direction, but more
layers should be added, and we point out some of the recent attempts at
hypermaps.

We propose that the main advantage arising from such a shift comes from the
ability to use artificial intelligence to process the information in the
hypermap to extract knowledge otherwise not available to the robot. This opens
interesting research opportunities both when considering how to perform this
knowledge extraction itself and for the benefit this formulation could bring to
high-level task planning.

%%%%%%%%%%%%%%%%%%%%%%%%%%%%%%%%%%%%%%%%%%%%%%%%%%%%%%%%%%%%%%%%%%%%%%%%%%%%%%%%

%\newpage

\bibliographystyle{IEEEtran}
\bibliography{refs}

\end{document}